\documentclass[10pt,twocolumn,letterpaper]{article}

\usepackage{cvpr}              

\usepackage{pifont}
\usepackage{comment}
\usepackage{multirow}
\usepackage{color,colortbl, xcolor}
\usepackage{makecell}
\usepackage{stfloats}
\usepackage{gensymb}
\usepackage{wrapfig}

\definecolor{CBRed}{RGB}{204, 102, 119}
\definecolor{CBYellow}{RGB}{221, 204, 119}
\definecolor{CBBlue}{RGB}{136, 204, 238}
\definecolor{CBGreen}{RGB}{0, 158, 115}
\definecolor{CBMagenta}{RGB}{170, 68, 153}
\definecolor{CBOrange}{RGB}{230, 159, 0}

\newcommand{\mypara}[1]{\vspace{3pt}\noindent\textbf{#1}}

\newcommand{\OURS}{MeshArt}

\def\ObjectPart{\mathcal{P}}
\def\ObjectStructure{\mathcal{S}}
\def\PartStructureInfo{\mathbf{s}}
\def\PartBoundingBox{\mathbf{b}}
\def\PartSemantics{\mathbf{l}}
\def\PartGeometry{\mathbf{g}}
\def\PartArticulationProp{\mathbf{a}}
\def\PartJointType{\mathbf{j}}
\def\PartJointExistence{\mathbf{f}}
\def\PartJointLocation{\mathbf{p}}
\def\PartJointOrientation{\mathbf{o}}

\def\StructureMesh{\mathcal{M}_s}
\def\StructureEmbeddings{\mathcal{Z}_s}
\def\StructureTokens{\mathcal{T}_s}

\def\PartMesh{\mathcal{M}_g}
\def\PartFaceEmbeddings{\mathcal{Z}_g}
\def\PartFaceTokens{\mathcal{T}_g}
\def\JunctionFaceProbability{p}

\def\GeometryEncoder{\mathbf{E}_g}
\def\GeometryDecoder{\mathbf{D}_g}

\def\StructureEncoder{\mathbf{E}_s}
\def\StructureDecoder{\mathbf{D}_s}
\def\StructureCodebook{\mathcal{C}_s}
\def\GeometryCodebook{\mathcal{C}_g}
\def\GeometryTransformer{\Phi_g}
\def\StructureTransformer{\Phi_s}
\definecolor{tabbestcolor}{rgb}{0.785, 0.851, 0.969}

\newcommand{\myparagraph}[1]{\vspace{1mm}\noindent\textbf{#1}~}

\definecolor{cvprblue}{rgb}{0.21,0.49,0.74}
\usepackage[pagebackref,breaklinks,colorlinks,citecolor=cvprblue]{hyperref}

\def\paperID{***} 
\def\confName{***}
\def\confYear{***}

\title{\OURS: Generating Articulated Meshes with Structure-Guided Transformers}

\author{
Daoyi Gao$^{1}$ \hspace{1cm}
Yawar Siddiqui$^{1,2}$ \hspace{1cm}
Lei Li$^{1}$ \hspace{1cm}
Angela Dai$^{1}$ \\
$^{1}$Technical University of Munich \hspace{1cm} $^{2}$Meta
}

\begin{document}




\begin{figure}
\twocolumn[{%
\renewcommand\twocolumn[1][]{#1}%
\maketitle
\begin{center}
    \centering
    \vspace{-0.6cm}
    \includegraphics[width=0.98\textwidth]{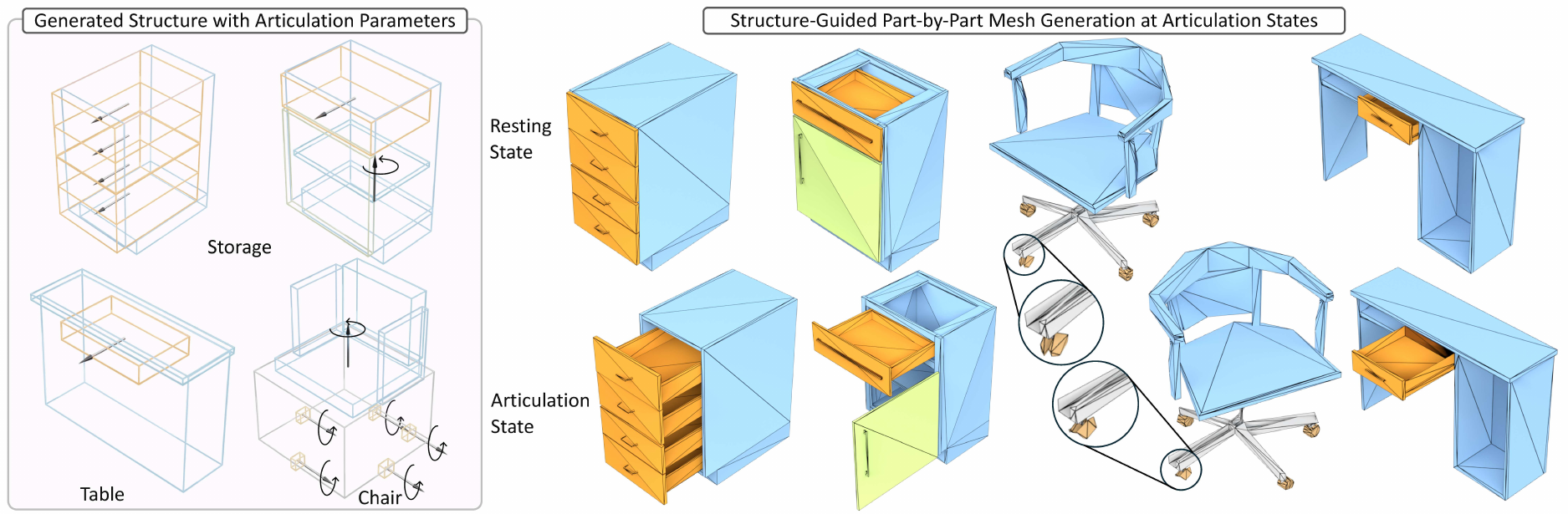}
    \vspace{-0.3cm}
    \caption{
    \OURS{} creates articulated 3D objects hierarchically, part-by-part with transformers by first generating high-level object structures endowed with articulations (\textbf{left}) and then predicting triangle mesh faces for each part (\textbf{right}) with structure and local connectivity guidance. Our mesh generations have clean and compact triangulations and can be animated at different articulation states.
    }
    \label{figure:teaser}
\end{center}%
}]
\end{figure}

\begin{abstract}
Articulated 3D object generation is fundamental for creating realistic, functional, and interactable virtual assets which are not simply static. 
We introduce \OURS{}, a hierarchical transformer-based approach to generate articulated 3D meshes with clean, compact geometry, reminiscent of human-crafted 3D models. 
We approach articulated mesh generation in a part-by-part fashion across two stages.
First, we generate a high-level articulation-aware object structure; then, based on this structural information, we synthesize each part's mesh faces. 
Key to our approach is modeling both articulation structures and part meshes as sequences of quantized triangle embeddings, leading to a unified hierarchical framework with transformers for autoregressive generation.
Object part structures are first generated as their bounding primitives and articulation modes; a second transformer, guided by these articulation structures, then generates each part's mesh triangles.
To ensure coherency among generated parts, we introduce structure-guided conditioning that also incorporates local  part mesh connectivity.
\OURS{} shows significant improvements over state of the art, with 57.1\% improvement in structure coverage and a 209-point improvement in mesh generation FID.

\end{abstract}  
\section{Introduction}
\label{sec:intro}

Articulated objects are ubiquitous in real-world environments -- for instance, cabinet doors are openable, office chairs typically swivel, car wheels and doors rotate and move.
These articulations are generally tied to the core functionality of these objects (\eg{}, storage for cabinets, ease of movement for chairs, movement of cars), and are thus key to representing realistic, interactable objects.
Such objects are already widely represented in computer graphics applications as triangle meshes, carefully crafted by skilled 3D artists.
This enables high visual quality while maintaining a compact representation that readily supports the articulated motion of various object parts.

Recent advances in generative modeling for vision and graphics have produced remarkable progress in generating static 3D objects~\cite{siddiqui2024meshgpt,cheng2023sdfusion,erkocc2023hyperdiffusion,shue20233d,zhou20213d,siddiqui2024assetgen} as well as complex static 3D scenes~\cite{wu2024blockfusion,roessle2024l3dg,meng2024lt3sd}, leveraging various 3D representations, from voxels and point clouds to neural fields and even 3D meshes.
However, generating articulated objects with dynamic, functional parts remains an underexplored challenge.
The primary difficulties lie in not only modeling possible functional part motions but also generating clean and compact part geometries that respect the articulation structures.
In particular, generating articulated objects as triangle meshes enables natural support of changing topological structure (e.g., open vs. closed doors in a shelf) while maintaining a compact representation similar to human-crafted objects.
Besides, current 3D datasets with object part and articulation joint annotations remain relatively limited in quantity for data-driven learning.

To address these challenges, we introduce \OURS{}\footnote{Project page: \href{https://daoyig.github.io/Mesh_Art/}{daoyig.github.io/Mesh\_Art/}}, a hierarchical transformer-based model that generates articulated 3D meshes with sharp details and an efficient triangle mesh representation, resembling the compactness of artist-created 3D models.
Our key insight is to decompose articulated meshes into high-level object structures and low-level part geometries: high-level structures are represented by part bounding primitives endowed with articulation properties, while part geometries are generated based on both articulation structures and local mesh connectivity.
To enable effective training of \OURS{}, we enhance PartNet~\cite{mo2019partnet}, the largest 3D object dataset with part labels, by adding articulation joint annotations to the table, chair, and storage categories, increasing the number of articulated objects $6\times$ compared to existing articulation datasets~\cite{geng2023gapartnet, xiang2020sapien}.

Our approach leverages a direct sequence generation approach with transformers to synthesize articulated objects hierarchically, part by part.
We first predict the object structure as a sequence of part bounding boxes, where each box encodes part semantics, latent geometric features, and articulation properties.
Using the predicted structure, we then generate the mesh geometry for each part as a sequence of triangles, conditioned on the structure and local connectivity from previously generated parts.
To unify this hierarchical generation process, we also parameterize part bounding boxes as triangle meshes; this unified representation enables more effective generation of coherent articulated meshes. 

At the object structure level, we learn a vocabulary of object structures by encoding triangulated part bounding boxes into quantized triangle embeddings, and then train a decoder-only transformer to autoregressively predict a triangle sequence representing the part bounding primitives.
At the part geometry level, we learn a separate codebook that encodes both part mesh triangles and their likelihood of serving as junctions between parts. We then train another transformer to generate the triangle sequence for each part mesh, conditioned on the object structure embeddings and junction faces from neighboring parts, ensuring smooth transitions between parts.
Experiments across three categories of our articulated PartNet dataset demonstrate that our method significantly improves both the diversity of articulation structures and the quality of generated 3D meshes compared to state-of-the-art methods, achieving an average improvement of 57.1\% in structure coverage and a 209-point improvement in mesh generation FID scores.

In summary, our contributions are: 

\begin{itemize}
    \item We propose \OURS{}, a novel hierarchical approach for generating articulated 3D objects, part by part to create compact triangle meshes. 
    Our approach unifies coarse articulation-aware structure generation with structure-guided part mesh generation, treating both as triangle sequence prediction tasks.
    \item We introduce a conditioning mechanism that integrates structure and local geometry connections to ensure smooth transitions during part-by-part mesh generation.
    \item We augment PartNet with joint annotations on 3 major categories, expanding the articulated object dataset by over 6 times.
\end{itemize}

\section{Related Work}
\label{sec:related}

\mypara{Mesh Generation.} 
Several methods have been proposed to directly generate mesh representation, due to their compact nature more similar to human-created 3D models.
Earlier approaches such as Scan2Mesh~\cite{dai2019scan2mesh} proposed to generate mesh structures as graphs, but were limited to small meshes with few vertices.
PolyGen~\cite{nash2020polygen} used two autoregressive transformers to model vertex and face distributions separately, while Polydiff~\cite{alliegro2023polydiff} applied diffusion to generate triangle soups. MeshGPT~\cite{siddiqui2024meshgpt} tokenized meshes via a GNN-based encoder and learned mesh token generation with a GPT-style transformer. PivotMesh~\cite{weng2024pivotmesh} improved generation by guiding it with pivot vertices to define coarse object shapes.
More recent approaches, including MeshAnythingV2~\cite{chen2024meshanythingv2} and EdgeRunner~\cite{tang2024edgerunner}, tokenized meshes by forming face sequences from adjacent vertices. MeshXL~\cite{chen2024meshxl} introduced spatial coordinate embeddings, while Meshtron~\cite{hao2024meshtron} employed an hourglass transformer for high-resolution mesh generation.

In contrast to these advances in static mesh generation, our work focuses on generating functional articulated objects by disentangling structure from geometry with hierarchical autoregressive models.

\mypara{Structured Object Generation.}
Structure-aware generative models for 3D objects have focused on modeling object parts and their relationships~\cite{li2017grass, wang2011symmetry}. 
SAGNet~\cite{wu2019sagnet} learns part geometry and pairwise relations within a shared latent space, while StructureNet~\cite{mo2019structurenet} represents objects as n-ary graphs, using a hierarchical network to encode both aspects.
SDM-NET~\cite{gao2019sdm} models part geometry with deformable mesh boxes and encodes global structure in a VAE. DSG-Net~\cite{yang2020dsm} learns separate latent space for geometry and structure, achieving fine-grained control during generation.

While these methods effectively generate coherent 3D shapes with structural relations, they do not model articulation. In contrast, we encode articulation properties at the object structure level, enabling the creation of compact meshes with functional part motions.

\mypara{Articulated Object Modeling.}
Recent advances in articulated 3D object modeling focus on reconstruction or generation with geometric and motion properties. Ditto~\cite{jiang2022ditto} reconstructs articulated objects from point clouds as implicit representations, capturing occupancy, part types, and joint parameters. Similarly, ~\cite{carto23} and ~\cite{rearticulable23} use implicit neural fields to reconstruct articulation from stereo RGB and point cloud videos. PARIS~\cite{liu2023paris} disentangles static and movable parts from two articulation states, predicting both geometry and motion. Building on this, \cite{weng2024neural} introduces explicit point-level correspondence between articulated states, enabling more complex object modeling.

NAP~\cite{lei2023nap} models 3D objects with an articulation tree, decoding final geometry via implicit fields or part retrieval. CAGE~\cite{liu2024cage} employs diffusion to generate articulation abstractions, relying on part retrieval for final output. However, retrieval-based approaches~\cite{zou20173d, wu2020pq, sung2017complementme} can bring inconsistencies in part geometry and overall structure.
In contrast, our approach generates structure and geometry cohesively in a unified triangle mesh representation, improving efficiency and robustness in articulated object modeling.

\section{Method}
\label{sec:method}

Our goal is to generate articulated 3D objects in a compact triangle mesh representation with sharp geometric details. We formulate articulated mesh generation in a hierarchical, part-by-part fashion and propose \OURS{}, a transformer-based approach that generates both part-level object structures and part mesh geometry as sequences of triangles.

To capture high-level object structures (\cref{sec:structure_generation}), we decompose a 3D object into functional parts $\{\ObjectPart^{i}\}$ and represent the object structure as $\ObjectStructure = \{\PartStructureInfo^{i}\}$.
Each $\PartStructureInfo^{i}$ encodes the bounding box, articulation properties, part semantics, and latent geometry features of part $\ObjectPart^{i}$.
This part-level abstraction enables efficient joint modeling of parts and articulation modes, providing global guidance for articulated mesh generation.
To formulate structure generation as a sequence prediction task, we represent each part's bounding box as a triangle mesh and learn a structure codebook $\StructureCodebook$ to encode $\ObjectStructure$ as a sequence of quantized triangle embeddings.
A structure transformer $\StructureTransformer$ is trained to predict the next embedding index in this sequence.
Once trained, $\StructureTransformer$ can be autoregressively sampled to generate articulated object structures.

Next, we generate part geometry for $\ObjectPart^{i}$ as a sequence of mesh triangles, guided by the global object structure $\ObjectStructure$ (\cref{sec:part_generation}).
To further capture local geometry and connectivity between parts, we encode the mesh triangles of $\ObjectPart^{i}$, along with their likelihood of being junction faces connecting different parts, as a quantized embedding sequence over a separately learned geometry codebook $\GeometryCodebook$.
This junction injection ensures smooth transitions between adjacent parts.
A geometry transformer $\GeometryTransformer$ then predicts the next triangle face embedding index, conditioned on $\ObjectStructure$ and junction embeddings from neighboring parts.
Finally, we decode the generated embeddings into a clean, compact part mesh.

\begin{figure}[tp]
    \centering
    \includegraphics[width=\columnwidth]{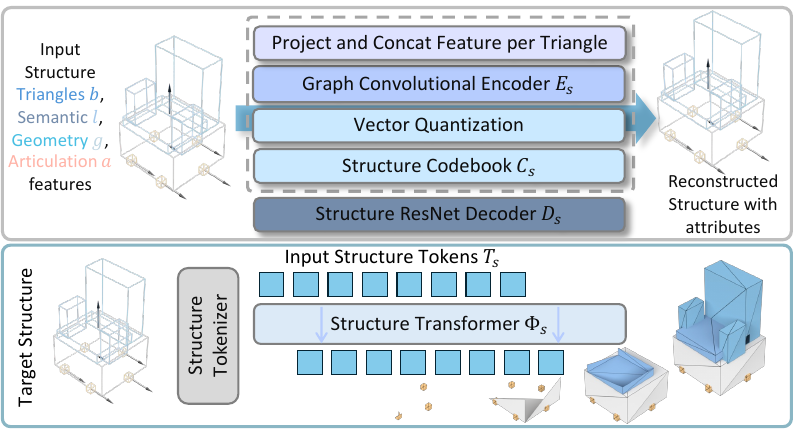}  
    \vspace{-0.7cm}
    \caption{We model high-level object structures with articulations using a VQ-VAE for structure encoding (\textbf{top}) and a decoder-only transformer for autoregressive structure sequence generation (\textbf{bottom}).
    The encoder $\StructureEncoder$ first encodes triangulated part bounding boxes, along with semantics, coarse geometry, and articulation features, into a structure token sequence $\StructureTokens$ over a learned codebook $\StructureCodebook$. The structure transformer $\StructureTransformer$ then learns to predict the next token in $\StructureTokens$, which is subsequently decoded by the decoder $\StructureDecoder$ to generate the object structure.
    }
    \vspace{-0.4cm}
    \label{fig:pipeline_structure}
\end{figure}

\subsection{Articulation-Aware Structure Generation}
\label{sec:structure_generation}

For an articulated object with functional parts $\{\ObjectPart^{i}\}$, we represent its high-level structure as $\ObjectStructure = \{\PartStructureInfo^{i}\}$, capturing each part's spatial location, articulation motion, semantic label, and coarse geometry guidance.
Encoding articulation at the structure level abstracts away part geometry details, enabling efficient modeling of shared articulation modes across objects.
Articulation modeling requires distinguishing between various articulated and static parts, and defining articulation properties, including joint types, locations, and orientations, as detailed below.

\mypara{Structure Parameterization.}
For each part $\ObjectPart^{i}$, we encode its bounding box $\PartBoundingBox^{i}$, articulation properties $\PartArticulationProp^{i}$, semantic label $\PartSemantics^{i}$, and latent geometry $\PartGeometry^{i}$ as  
\begin{equation}
    \PartStructureInfo^{i} = (\PartBoundingBox^{i}, \PartArticulationProp^{i}, \PartSemantics^{i}, \PartGeometry^{i}).
\end{equation}

The \emph{part bounding box} $\PartBoundingBox^{i}$ is represented as an axis-aligned bounding box (AABB) in the object's canonical resting state (\eg, doors and drawers are closed). For efficient sequence modeling and generation, as well as effective conditioning for the mesh surface generation, we represent these bounding boxes as triangle meshes. 

\emph{Articulation properties} $\PartArticulationProp^{i} = (\PartJointType^{i}, \PartJointExistence^{i}, \PartJointOrientation^{i}, \PartJointLocation^{i})$ include joint type $\PartJointType^{i}$, existence $\PartJointExistence^{i}$, orientation $\PartJointOrientation^{i}$, and location $\PartJointLocation^{i}$.
Three joint types are considered: \textit{fixed}, \textit{revolute}, and \textit{prismatic}.
The joint orientation is represented as a 3D unit vector.
Prismatic joint locations are canonicalized to the object's origin.
The range for revolute joints is set to $90\degree$, and the range for prismatic joints is determined by the object's base extent.

The \emph{part semantic label} $\PartSemantics^{i}$ is encoded as a feature vector using the CLIP text encoder~\cite{radford2021learning}.

\mypara{Learning Quantized Structure Embeddings.}
To facilitate autoregressive generation of articulated object structures with transformers, we encode the structure information $\ObjectStructure$ as a sequence of quantized triangle embeddings, as shown in \cref{fig:pipeline_structure}-top.
We adopt a bottom-up sequence ordering \cite{siddiqui2024meshgpt,nash2020polygen}, arranging the $N$ part bounding box meshes $\{\PartBoundingBox^{i}\}$ as a sequence of triangles:
\begin{equation}
    \StructureMesh := (f^{1}_{1}, \ldots, f^{1}_{12}, \ldots, f^{i}_{j}, \ldots, f^{N}_{12}),
\end{equation}
where $f^{i}_{j}$ denotes the $j$-th triangle face of the $i$-th part bounding box mesh.
Bounding boxes are sorted from bottom to top based on their lowest face locations. For each box mesh, vertices are sorted from lowest to highest in $z$-$y$-$x$ order (vertical axis $z$), and the 12 triangle faces are then ordered by their lowest vertex, the next lowest, and so on.

Similar to MeshGPT~\cite{siddiqui2024meshgpt}, we then train a Vector Quantized-Variational Autoencoder (VQ-VAE)~\cite{van2017neural}; instead of learning a codebook encoding mesh triangles, we learn a structure codebook $\StructureCodebook$ encoding a triangle-based representation of the coarse structure.
The structure encoder $\StructureEncoder$ employs graph convolutions~\cite{hamilton2017inductive} by treating the triangles in $\StructureMesh$ as nodes.
Structure features from $\ObjectStructure$ are independently projected and then concatenated with positionally encoded triangle coordinates as input features.
The structure encoder $\StructureEncoder$ extracts an articulation-aware feature vector for each structure triangle as
\begin{equation}
    \StructureEmbeddings = (\mathbf{z}^{1}_{1}, \ldots, \mathbf{z}^{i}_{j}, \ldots, \mathbf{z}^{N}_{12}) = \StructureEncoder(\StructureMesh, \ObjectStructure),
\end{equation}
where $\mathbf{z}^{i}_{j}$ denotes the feature vector for $f^{i}_{j}$.
 
At the bottleneck, a residual vector quantization (RQ) module~\cite{martinez2014stacked} assigns $D$ codes to each structure triangle $f^{i}_{j}$, based on the learned embedding $\mathbf{z}^{i}_{j}$ and codebook $\StructureCodebook$, resulting in a structure token sequence $\StructureTokens$ as
\begin{equation}
    \StructureTokens = (t^{1}_{1}, \ldots, t^{i}_{j}, \ldots, t^{N}_{12}) = \text{RQ}(\StructureEmbeddings; \StructureCodebook, D),
\end{equation}
where $t^{i}_{j}$ consists of $D$ embedding indices.
A 1D-ResNet~\cite{he2016deep} decoder $\StructureDecoder$ then decodes the quantized face embedding sequence to reconstruct the structure $\ObjectStructure$.

During training, to supervise the reconstruction of $\ObjectStructure$, we apply a cross-entropy loss to predict the quantized locations of bounding box triangles $\PartBoundingBox^{i}$, joint type $\PartJointType^{i}$, joint existence $\PartJointExistence^{i}$, and joint location $\PartJointLocation^{i}$.
Predicting locations as discretized coordinates helps stabilize training.
We use an $\ell_2$ loss for regressing the joint orientation $\PartJointOrientation^{i}$, semantic feature $\PartSemantics^{i}$, and geometry feature $\PartGeometry^{i}$.
We refer to the supplemental for more training details on the structure VQ-VAE.

\mypara{Structure Generation with Transformers.}
Given the learned structure codebook $\StructureCodebook$, we use a GPT-style decoder-only transformer $\StructureTransformer$ to predict the structure token sequence $\StructureTokens$, which is then decoded into an object structure $\ObjectStructure$ by $\StructureDecoder$ (\cref{fig:pipeline_structure}-bottom).
The structure transformer $\StructureTransformer$ is trained with a cross-entropy loss for next-token index prediction. Once trained, we use beam sampling to generate articulated structures unconditionally from $\StructureTransformer$.

\subsection{Structure-Guided Part Mesh Generation}
\label{sec:part_generation}

\begin{figure*}[ht]
    \centering
    \includegraphics[width=\textwidth]{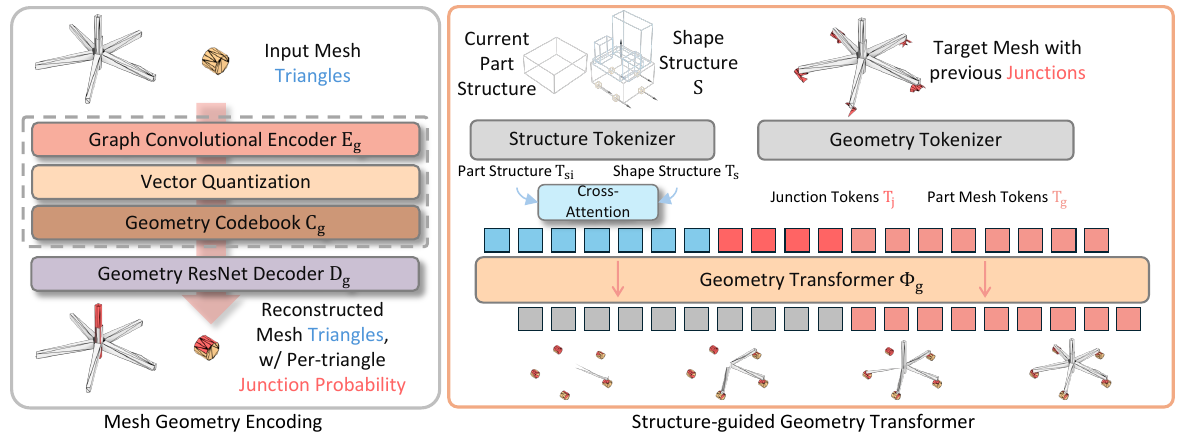}  
    \vspace{-0.7cm}
    \caption{We generate mesh triangles for a 3D object part by part, guided by the high-level object structure $\ObjectStructure$ and junction connections to previously generated part meshes.
    \textbf{Left}: The encoder $\GeometryEncoder$ encodes triangles of each part mesh into a sequence of quantized triangle embeddings $\PartFaceTokens$, using a separately learned geometry codebook $\GeometryCodebook$. The decoder $\GeometryDecoder$ then reconstructs the mesh triangles from $\PartFaceTokens$ and predicts a junction probability for each face connecting to other parts.
    \textbf{Right}: The geometry transformer is conditioned on the current part structure, the overall object structure $\StructureTokens$, and junction face tokens to predict the triangle tokens $\PartFaceTokens$ for generating the current part mesh.}
    \vspace{-0.5cm}
    \label{fig:pipeline_geometry}
\end{figure*}

Next, we generate mesh triangles for each part $\ObjectPart^{i}$, conditioned on the high-level object structure $\ObjectStructure$ using a geometry transformer $\GeometryTransformer$ over a separately learned triangle codebook $\GeometryCodebook$, as illustrated in \cref{fig:pipeline_geometry}.
Since we generate articulated object meshes part by part, we must ensure smooth connections to adjacent part meshes already generated. To address this, we introduce a junction face mechanism to guide triangle generation for a coherent 3D object mesh.

\begin{wrapfigure}[8]{r}{0.28\columnwidth}
    \hspace{-0.4cm}
    \includegraphics[width=0.28\columnwidth]{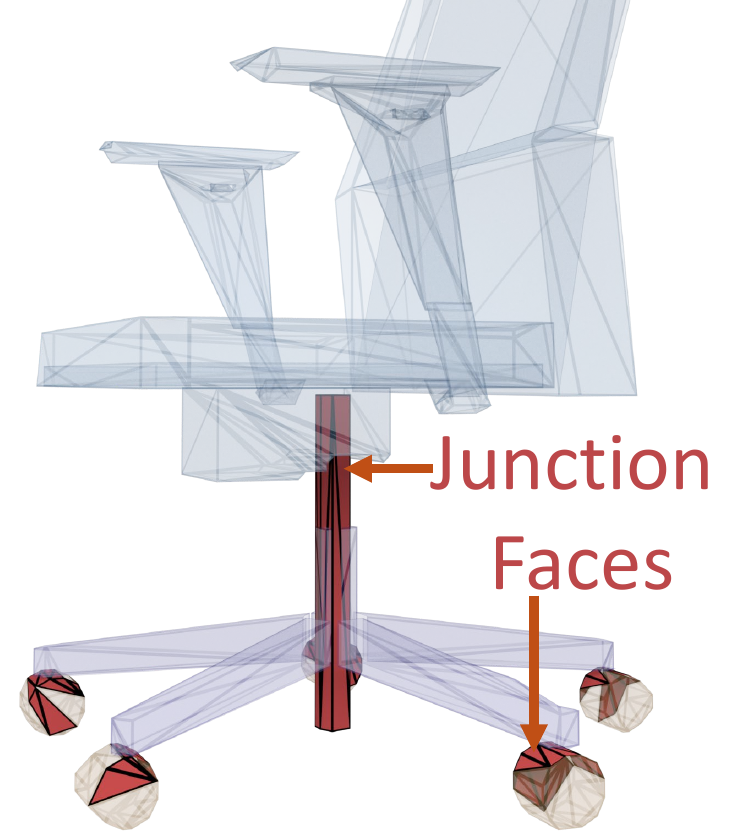}
    \vspace*{-1.5\baselineskip}
\end{wrapfigure}

\textit{Junction faces} are triangles that are spatially adjacent to different parts, providing crucial local geometry and connectivity cues for generating mesh triangles of a new part. 
For example, in the inset figure, junction faces of the chair wheels strongly indicate the chair base shape and connection points. 
We thus incorpoarate junction face information into the geometry codebook $\GeometryCodebook$ and condition the geometry transformer $\GeometryTransformer$ additionally on these junction faces to ensure structural continuity, as detailed below.

\mypara{Junction-aware Triangle Encoding.}
We first encode each part mesh $\ObjectPart^{i}$ as a sequence of quantized triangle embeddings using the geometry codebook $\GeometryCodebook$ (\cref{fig:pipeline_geometry}-left), similar to the structure embedding in \cref{sec:structure_generation}.

Triangles in $\ObjectPart^{i}$ are sorted bottom-up in a  sequence 
$\PartMesh^{i} = (f^{i}_{1}, f^{i}_{2}, \ldots, f^{i}_{N_i})$.
A graph convolutional~\cite{hamilton2017inductive} geometry encoder $\GeometryEncoder$ extracts informative geometric features $\PartFaceEmbeddings^{i}$ for the triangles as:
$\PartFaceEmbeddings^{i} = \GeometryEncoder(\PartMesh^{i})$.
An RQ module then quantizes these features, assigning $D$ codes per triangle based on the learned codebook $\GeometryCodebook$ as
$\PartFaceTokens^{i} = \text{RQ}(\PartFaceEmbeddings^{i}; \GeometryCodebook, D)$.

We employ a 1D ResNet decoder with two prediction heads.
One head reconstructs the  triangle coordinates of $\ObjectPart^{i}$ from the quantized embeddings, supervised by a cross-entropy loss.
The other head predicts the probability $\JunctionFaceProbability^{i}_{k}$ for each face $f^{i}_{k} \in \PartMesh^{i}$ to be a junction face connecting to other parts, thereby injecting junction information into $\GeometryCodebook$.
To supervise junction predictions, we identify the ground-truth junction faces by calculating the distance between each triangle in $\ObjectPart^{i}$ and neighboring parts, labeling as junction faces if their distances fall below a predefined threshold.

\mypara{Part Mesh Generation with Transformers.}
We train the geometry transformer $\GeometryTransformer$ to predict the triangle token sequence $\PartFaceTokens^{i}$, guided by the object structure $\ObjectStructure$ and junction face tokens from neighboring parts (\cref{fig:pipeline_geometry}-right).

The input sequence for $\GeometryTransformer$ is constructed by placing the structure and junction tokens at the beginning, followed by the triangle tokens $\PartFaceTokens^{i}$.
Concretely, for the current part mesh $\ObjectPart^{i}$, its corresponding structure tokens $ (t^{i}_{1}, \ldots, t^{i}_{12}) \subset \StructureTokens$ are extracted and cross-attended to the overall object structure $\StructureTokens$, before being added to the start of the sequence.
Next, the junction faces from other parts are appended to the sequence before triangle tokens $\PartFaceTokens^{i}$, informing smooth transitions and connectivity between neighboring parts.

The complete sequence is then fed into $\GeometryTransformer$, which has a stack of multi-head self-attention layers, to predict the next triangle token in the sequence.
We apply a cross-entropy loss to the predicted triangle tokens $\PartFaceTokens^{i}$, while the injected structure and junction tokens serve solely as conditioning.

Due to the varying length of the condition sequence, depending on the number of junction faces, standard index-based positional embeddings would result in inconsistent embeddings for the start token of the mesh sequence $\PartFaceTokens^{i}$. We employ a flexible positional embedding scheme to address this: we fix the positional embeddings starting from $\PartFaceTokens^{i}$, while applying a constant positional embedding vector for the preceding condition sequence.

\subsection{Hierarchical Articulated Mesh Sampling}
At inference time, we first sample the structure transformer  $\StructureTransformer$ unconditionally to generate an object structure sequence $\ObjectStructure$, which is then decoded by the structure decoder $\StructureDecoder$
into part bounding boxes with articulation properties.

Next, for each part $\ObjectPart^{i}$, we generate its triangle sequence $\PartFaceTokens^{i}$ by sampling the geometry transformer $\GeometryTransformer$, conditioned on object structure $\ObjectStructure$ and junction faces from nearby part meshes. 
We cache all the triangles predicted as junctions by $\GeometryDecoder$, then retrieve the junction faces from the cache based on their proximity to the current structure bounding box $\PartBoundingBox^{i}$.

The predicted triangle tokens $\PartFaceTokens^{i}$ are then decoded into mesh triangles by the geometry decoder $\GeometryDecoder$.
Finally, the object mesh is obtained by merging close vertices of the generated triangles.
This articulated mesh can be transformed into different articulation states based on the joint predictions.

\section{Articulated PartNet Dataset Annotations}
\label{sec:dataset_annoatation}

Previous methods for generating articulated objects~\cite{lei2023nap, liu2024cage} primarily relied on the PartNet-Mobility dataset~\cite{xiang2020sapien}, which comprises a small number of objects (c.f. \cref{tab:data_annotation}) with limited diversity in articulated structures. We thus instead augment the PartNet dataset~\cite{mo2019partnet} across 3 major categories by annotating articulation information for functional parts such as cabinet doors, drawers, and wheels. As shown in \cref{tab:data_annotation}, our annotations increase the number of articulated parts by more than $6\times$ compared to PartNet-Mobility.

To ensure accurate annotations of joint locations and orientations, we 
1) manually corrected inconsistencies in mesh orientations and part groupings; 
2) automatically annotated prismatic joints, aligning all joints to the object space origin;
3) generated hypotheses for revolute joints and manually selected the most plausible options through an interactive interface; 
and 4) manually verified and adjusted joints based on renderings of articulated part motions. 
Overall, we dedicated over 150 hours to constructing this PartNet extension.
We refer to the supplemental for further details.

{
\begin{table}[t]
  \begin{center}
    \small
    \resizebox{\linewidth}{!}{
      \begin{tabular}{lrrr}
        \toprule
        Categories        & Storage Furniture & Table & Chair \\
        \midrule
        PartNet-Mobility~\cite{xiang2020sapien}        & 346                           & 101                & 81                   \\
        Ours (Articulated PartNet)                       & \textbf{1786}                 & \textbf{1454}                  & \textbf{372}                    \\
        \bottomrule
      \end{tabular}}
      \vspace{-0.2cm}
    \caption{Our annotated articulated PartNet~\cite{mo2019partnet} significantly increases the number of articulated objects.}
    \vspace{-0.5cm}
    \label{tab:data_annotation}
  \end{center}
\end{table}
}


\section{Experiments}
\label{sec:experiments}

We present results using our annotated Articulated PartNet. 

\mypara{Implementation.}
To learn the structure and geometry codebooks, our RQ layer has a depth of 2, yielding $D=6$ embeddings per face, each with dimension 192. 
For the triangle location prediction, both structure and geometry decoders predict face coordinates across 128 classes, resulting in a discretization of space to $128^3$ possible values. 
The structure decoder also regresses the 768-dim part semantic features and 128-dim latent geometry features. 
For the structure transformer $\StructureTransformer$, we use a GPT2-small model with a context window of 4608, and for the geometry transformer $\GeometryTransformer$, we use a GPT2-medium model of the same window length.
We implemented our approach using PyTorch~\cite{NEURIPS2019_9015}.

\mypara{Training.} 
During training, we employ augmentation techniques including random shifts and random scaling to enhance the training mesh diversity. Similar to MeshGPT~\cite{siddiqui2024meshgpt}, we apply planar decimation to augment the shapes. To ensure that a part mesh fits into the geometry transformer's context window, we select the object parts with $<700$ faces (post-decimation) for training. 

We begin by training both the VQ-VAEs and transformer models across all 3 (chairs, tables, storage) categories and then fine-tune the transformers on each category individually.
We use a 90/10 training/testing split for each category. 
The structure VQ-VAE is trained on one A6000 GPU for 3 days, and the geometry VQ-VAE is trained on one A100 for 2 days until convergence.
Pretraining $\StructureTransformer$ takes 3 days on a single A100 GPU, followed by 1 day for fine-tuning on each category. 
Pretraining $\GeometryTransformer$ takes $\sim4$ days on four A100 GPUs, followed by 2 days per category for fine-tuning on a single A100.
We use the ADAM optimizer~\cite{kingma2014adam} with a learning rate of $1\times10^{-4}$ and a batch size of 64.

\mypara{Baselines.}
NAP~\cite{lei2023nap} is trained on our data on each category individually. CAGE~\cite{liu2024cage} requires an object part graph as input, so we extracted the graph conditions from our dataset and retrained CAGE with our data for all comparisons.

\subsection{Evaluation Metrics.} 
\noindent \textbf{Shape Quality.}
Following~\cite{lei2023nap, liu2024cage}, we use Instantiation Distance (ID) to assess mesh geometry at different articulation states and Abstract Instantiation Distance (AID) to evaluate object structure at these states.
As in~\cite{liu2024cage}, we synchronize articulation states of both generated and ground-truth meshes by sampling 10 evenly spaced articulation states between start and end positions. To evaluate ID and AID, we sample 2,048 surface points from each object mesh and its structure bounding box, respectively. We then compute Minimum Matching Distance (MMD), Coverage (COV), and 1-Nearest-Neighbor Accuracy (1-NNA). Lower MMD indicates better quality, higher COV reflects broader distribution coverage; for 1-NNA, 50\% is optimal.

\noindent \textbf{Perceptual Quality.}
While the metrics above assess geometric quality, they do not capture visual similarity to the real data distribution. To address this, we render generated and ground-truth meshes at the same articulation state from 8 viewpoints in Blender, using a metallic material to highlight geometric details. We then compute FID (Fréchet Inception Distance) and KID (Kernel Inception Distance), where lower scores indicate better performance.

\subsection{Results}

\begin{figure*}[ht]
    \centering
    \vspace{-0.3cm}
    \includegraphics[width=0.95\textwidth]{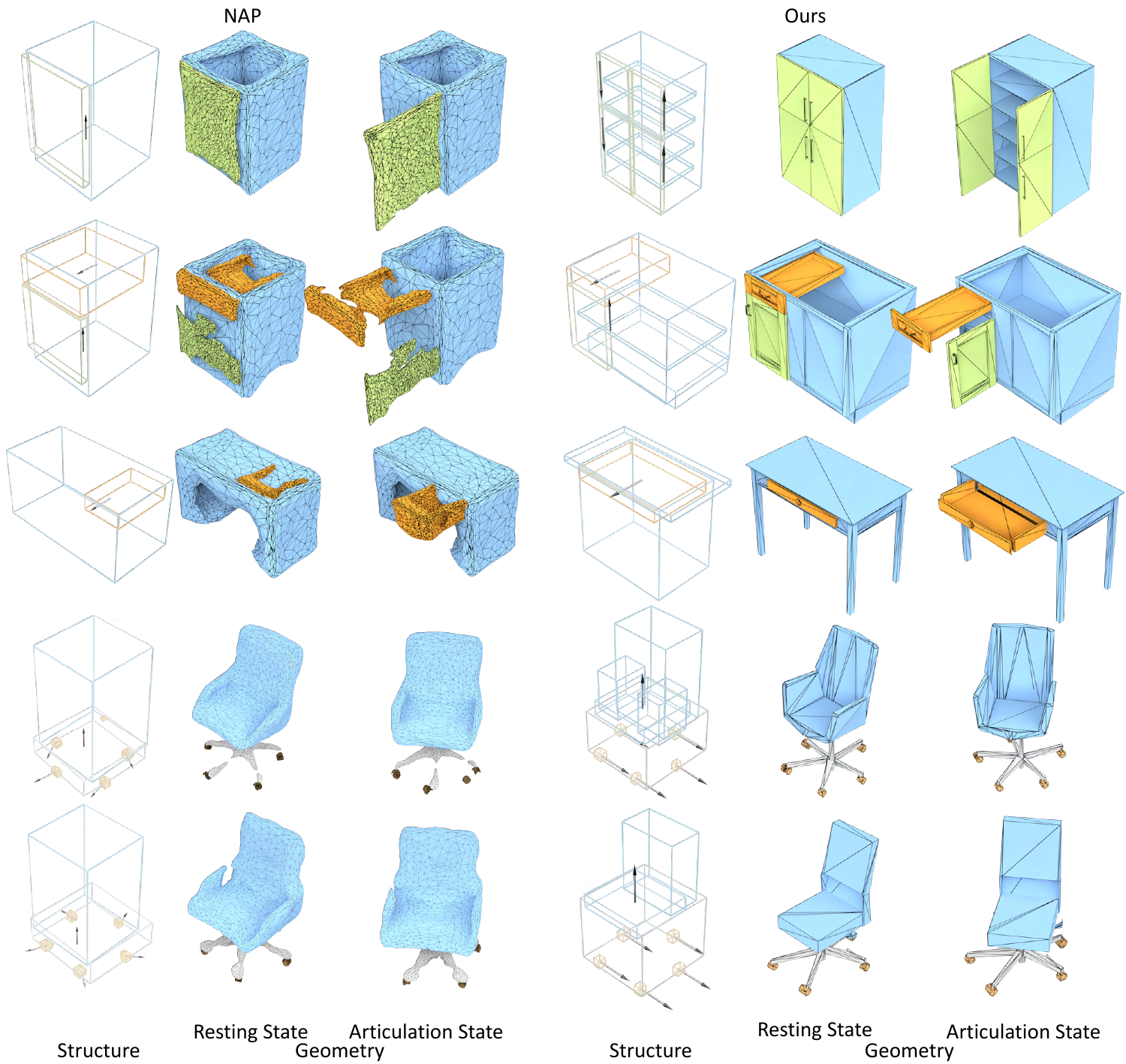}  
    \caption{
    Qualitative comparison across all categories, showing generated structures with joints, mesh geometry in resting and articulated states. Our approach generates compact meshes with sharp geometry and precise joints, enabling realistic part movements.
    }
    \vspace{-0.2cm}
    \label{fig:qualitative_comparison}
\end{figure*}
\begin{table}[tp]
  \begin{center}
    \small
    \resizebox{0.8\linewidth}{!}{
      \begin{tabular}{p{10mm}lrrrrr}
        \toprule
        \multirow{2}{*}{Class}                           & \multicolumn{1}{c}{\multirow{2}{*}{Method}}                  & \multicolumn{3}{c}{AID}                              \\
        \cmidrule(l){3-5} 

                                                         &                                                              & COV$\uparrow$ &   MMD$\downarrow$ & 1-NNA         \\
        \midrule

        \multirow{3}{*}{Chair}                           & NAP ~\cite{lei2023nap}                                       &    28.3       &      3.7       &       89.5     \\

                                                         & CAGE~\cite{liu2024cage}                                     &     {32.9}         &        {3.9}          &      {92.6}        \\

        \cmidrule{2-5}
                                                         & \OURS{}                                                      & \textbf{43.3}     & \textbf{3.6}       & \textbf{80.2}     \\
        \bottomrule

        \multirow{3}{*}{Table}                           & NAP ~\cite{lei2023nap}                                       &     21.1      &       3.0       &     92.3      \\

                                                         & CAGE~\cite{liu2024cage}                                     &      {25.9}      &       {3.9}       &         {88.4}   \\

        \cmidrule{2-5}
                                                         & \OURS{}                                                      & \textbf{40.2} & \textbf{2.3}    & \textbf{71.8} \\
        \bottomrule

        \multirow{3}{*}{\shortstack{Storage\\Furniture}} & NAP ~\cite{lei2023nap}                                       &     30.6      &      2.6     &     85.9    \\

                                                         & CAGE~\cite{liu2024cage}                                     &      {33.4}       &    {4.7}      &     {80.3}          \\

        \cmidrule{2-5}
                                                         & \OURS{}                                                      & \textbf{39.1} & \textbf{2.1}    & \textbf{77.1} \\
        \bottomrule
      \end{tabular}}
    \vspace{-2mm}
    \caption{Quantitative evaluation on articulated \emph{structure} generation. MMD values are multiplied by $10^3$. We outperform baselines in all 3 categories in structure generation diversity and realism. }
    \label{tab:articulation_structure}
    \vspace{-0.7cm}
  \end{center}
\end{table}

\begin{table}[b]
  \begin{center}
  \vspace{-0.3cm}
    \small
    \resizebox{\linewidth}{!}{
      \begin{tabular}{p{10mm}lrrrrrr}
        \toprule
        \multirow{2}{*}{Class}                           & \multicolumn{1}{c}{\multirow{2}{*}{Method}}                  & \multicolumn{5}{c}{ID}                                                                                                       \\
        \cmidrule(l){3-7}

                                                         &                                                              & COV$\uparrow$          & MMD$\downarrow$         & 1-NNA         & FID$\downarrow$ & KID$\downarrow$  \\
        \midrule

        \multirow{3}{*}{Chair}                           & NAP~\cite{lei2023nap}                                        &   37.1                 &       5.4               &    88.9        &     267.7            &      0.263         \\

        \cmidrule{2-7}
                                                         & \OURS{}                                                      & \textbf{44.0}              & \textbf{4.2}         & \textbf{73.2}  & \textbf{40.8}       & \textbf{0.008}   \\
        \bottomrule

        \multirow{3}{*}{Table}                           & NAP~\cite{lei2023nap}                                        &    27.9                &     6.7                 &     89.0       &    252.6     &        0.238             \\

        \cmidrule{2-7}
                                                         & \OURS{}                                                      & \textbf{36.9}          & \textbf{4.6}            & \textbf{78.6}     & \textbf{14.3}       & \textbf{0.002}  \\
        \bottomrule

        \multirow{3}{*}{\shortstack{Storage\\Furniture}} & NAP~\cite{lei2023nap}                                       &     33.3            &        4.0             &     94.1    &    170.6     &     0.167         \\

        \cmidrule{2-7}
                                                         & \OURS{}                                                     & \textbf{41.2}          & \textbf{3.1}            & \textbf{83.3} & \textbf{8.1}   & \textbf{0.002}  \\
        \bottomrule
      \end{tabular}}
    \vspace{-2mm}
    \caption{Quantitative evaluation on articulated \emph{mesh} generation. MMD values are multiplied by $10^3$. Our approach  outperforms baselines in both  shape and visual quality metrics.}
    \label{tab:articulation_mesh}
    \vspace{-0.5cm}
  \end{center}
\end{table}

\mypara{Structure Generation Quality.}
We evaluate how well our structure transformer covers the articulation structure distribution by unconditionally sampling our model and generating structures at least the same number as the train set. Using the generated joint information, we transform these structures across various articulation states and evaluate their AID scores, with results reported in Tab.~\ref{tab:articulation_structure}. As shown, our method consistently outperforms both baselines across all three categories.

NAP generates articulation trees where nodes store part geometry information and edges capture relative transformations and joint information using Pl\"ucker coordinates. 
It needs a post-processing step to extract a valid articulation tree and suffers from modeling intricate articulations, such as the joints are incorrectly predicted for the complex office chairs  shown in the last 2 rows of Fig.\ref{fig:qualitative_comparison}.
CAGE requires a category label and a connectivity graph as conditions to produce articulation abstractions. However, CAGE's generated structures exhibit repetitive patterns, resulting in lower coverage scores.
Our method achieves lower MMD and higher COV scores, demonstrating its ability to produce more realistic and diverse articulation structures.

\mypara{Articulated Mesh Generation.} 
A key objective of our work is to generate articulated meshes with clean, sharp geometry. As demonstrated in Fig.~\ref{fig:qualitative_comparison} and Tab.~\ref{tab:articulation_mesh}, our method consistently outperforms NAP~\cite{lei2023nap} in ID score and visual quality. NAP requires Marching Cubes~\cite{lorensen1998marching} to decode the final geometry from generated shape codes, resulting in over-smoothed and over-tessellated meshes. In contrast, our approach produces sharper and more compact meshes, effectively preserving sharp geometry details.

\begin{figure}[bt]
    \centering
    \vspace{-0.3cm}
    \includegraphics[width=0.9\columnwidth]{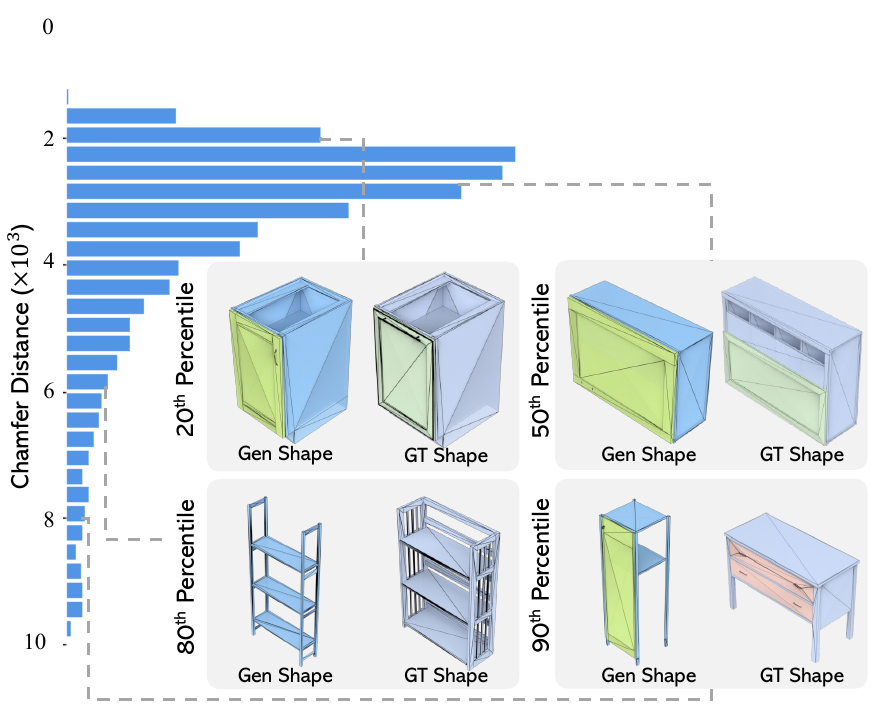}  
    \caption{Shape novelty analysis on storage category. We generate 1,000 shapes and plot their distances to the train set. The  generated shape (left) at the 50$^{\text{th}}$ percentile already differs from its closest match (right), showing that our method captures both the shape distribution (low CD) while generating novel shapes (high CD).}
    \label{fig:nn_eval}
\end{figure}

\mypara{Shape Novelty Analysis.}
We analyze the shape generation novelty following~\cite{weng2024pivotmesh, siddiqui2024meshgpt}. We generate 1,000 shapes and retrieve their nearest neighbors in the training set using Chamfer Distance (CD).  
As shown in Fig.~\ref{fig:nn_eval}, our method captures the training distribution well (indicated by low CD) while effectively generating novel shapes (with high CD).

\begin{figure}[t]
    \centering
    \includegraphics[width=0.95\columnwidth]{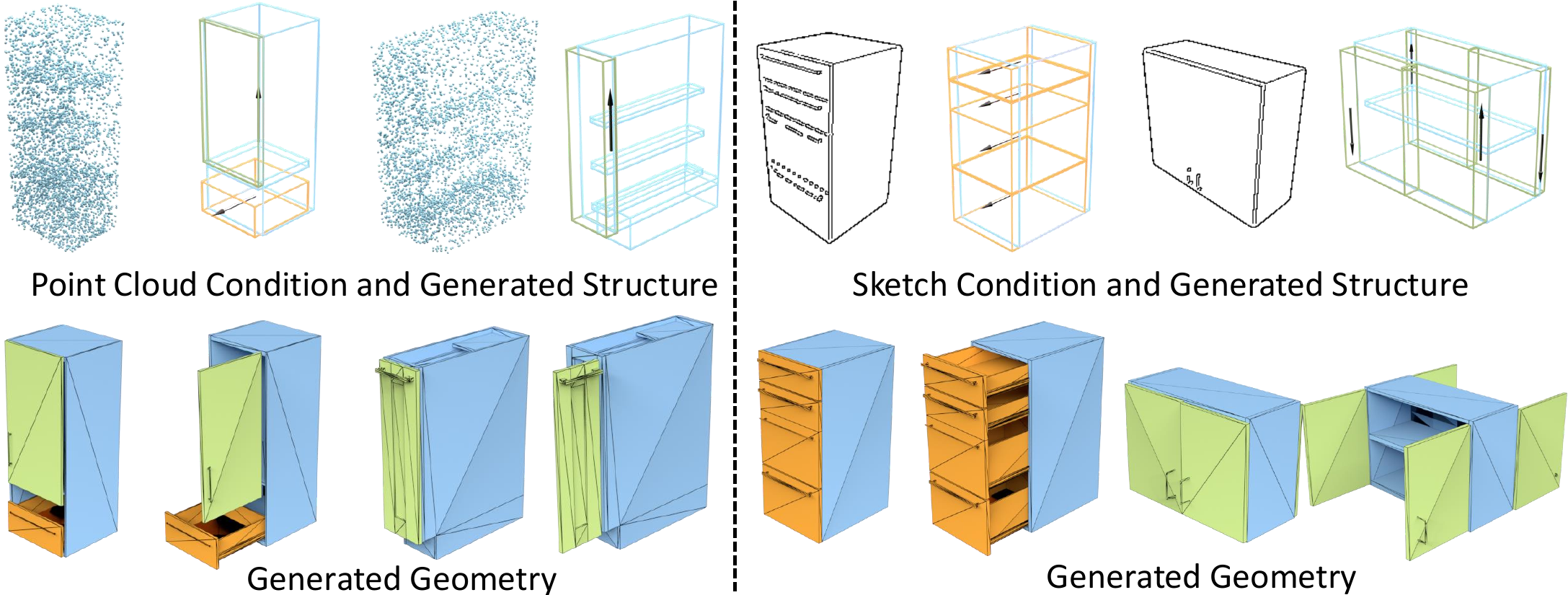}  
    \caption{Conditional Generation. Articulated structures and geometry are generated conditioned on point clouds or sketches.
    }
    \vspace{-0.2cm}
    \label{fig:condgen}
\end{figure}

\mypara{Conditional Generation.}
Our approach can support conditional generation by injecting condition tokens to the structure transformer $\StructureTransformer$ while keeping the geometry transformer $\GeometryTransformer$ fixed. As shown in Fig.~\ref{fig:condgen}, our approach generates articulated meshes from 3D point clouds and sketch images. More results can be found in the supplemental.
\subsection{Ablations}

\mypara{Are triangles a better representation for object structures?}
We consider using the bounding box vertices or the min/max corners as structure parameterizations, encoded with  PointNet~\cite{qi2017pointnet}. As shown in Tab.~\ref{tab:ablations_structure_parameterization}, our triangle parameterization significantly outperforms the baselines, enabling more diverse and realistic articulated structure generation.

\begin{figure}[t]
    \centering
    \includegraphics[width=\columnwidth]{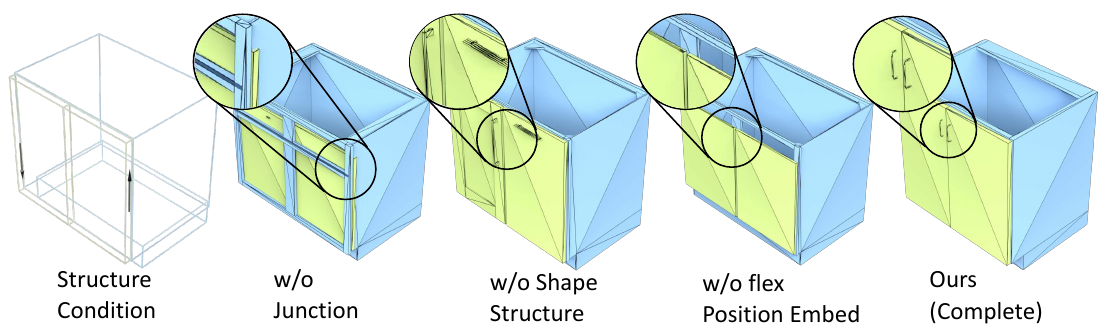}  
    \vspace{-0.3cm}
    \caption{Ablation visualization. Omitting junction guidance (w/o junction) results in part incoherence, while excluding global structure awareness (w/o Shape Structure) produces coherent shapes but lacks style consistency across parts. Without the flexible embedding scheme, the model tends to generate simpler shapes.}
    \vspace{-0.2cm}
    \label{fig:ablation}
\end{figure}

\mypara{Effectiveness of junction face conditioning.}
Tab.~\ref{tab:ablations} and Fig.~\ref{fig:ablation} shows the effect of our junction face conditioning, in comparison to conditioning only on shape structure information. This notably improves shape consistency and results in more plausible synthesis.

\mypara{Effect of shape structure conditioning.}
Fig.~\ref{fig:ablation} and  Tab.~\ref{tab:ablations} show that full shape structure conditioning enables more coherent shape generation, thus improving performance.

\mypara{Effect of flexible embedding.}
Without the flexible positional embedding, variation in the condition sequence length leads to inconsistent positional embedding features for the mesh part starting token. This inconsistency hinders learning of long-range dependencies, resulting in generating simpler shapes, as illustrated in Fig.~\ref{fig:ablation}.

{
\begin{table}[tbp]
  \begin{center}
    \small
    \resizebox{0.65\linewidth}{!}{
      \begin{tabular}{lrrr}
        \toprule
        \multirow{2}{*}{Method}                                & \multicolumn{3}{c}{AID}                              \\
        \cmidrule(l){2-4}
                               & COV$\uparrow$  & MMD$\downarrow$ & 1-NNA         \\
        \midrule
        Min/Max Bounds               &      36.3      &      4.6        &     84.0            \\
        Bbox Corners                 &      35.0      &      4.4      &      85.7       \\
        \midrule
        \OURS{}                      & \textbf{39.1} & \textbf{2.1}   & \textbf{77.1} \\
        \bottomrule
      \end{tabular}}
    \vspace{-2mm}
    \caption{{Ablation study on structure parameterizations (storage category). Using triangles as parameterization notably improves structure generation quality compared to baselines.}}
    \label{tab:ablations_structure_parameterization}
    \vspace{-0.3cm}
  \end{center}

\end{table}
}
{
\begin{table}[tp]
  \begin{center}
    \small
    \resizebox{\linewidth}{!}{
      \begin{tabular}{lrrrrr}
        \toprule
        Method                       & COV$\uparrow$  & MMD$\downarrow$ & 1-NNA          & FID$\downarrow$ & KID$\downarrow$  \\
        \midrule
        w/o Junction Condition       &      39.1      &       2.5       &    80.0        &        16.7     &      0.006        \\
        w/o Shape Structure Guidance &     39.7       & 2.1            & 79.7          &      9.6      &      0.004       \\
        w/o Flex Positional Embedding &     40.2       & 2.1            &   79.5        &       8.8     &      \textbf{0.003}       \\
        \cmidrule{1-6}
        \OURS{}                      & \textbf{40.9} & \textbf{2.0}   & \textbf{78.7} & \textbf{8.6}   & \textbf{0.003} \\
        \bottomrule
      \end{tabular}}
    \vspace{-2mm}
    \caption{Ablation study (storage category). Incorporating structure, local vicinity junctions, and a flexible positional embedding enables our model to generate more diverse and realistic shapes.}
    \label{tab:ablations}
    \vspace{-0.3cm}
    
  \end{center}

\end{table}
}

\mypara{Limitations.}
While \OURS{} shows strong potential in 3D articulated mesh generation, some limitations remain. 
Our transformer-based approach predicts codebook indices for structure and geometry, without full awareness of the decoded surface or articulation motion, 
making it challenging to enforce physical plausibility, as the codebook selection is non-differentiable.
Additionally, while our Articulated PartNet annotations significantly expand training data, its size remains limited for large-scale training.
Nevertheless, we see this dataset as a valuable step toward enabling compact articulated 3D mesh generation.

\section{Conclusion}
\label{sec:conclusion}

We introduced \OURS{}, a triangle-based hierarchical approach for generating articulated 3D meshes. We decompose objects into coarse part-based structures and detailed part mesh geometries, grounding articulations at the structure level. This enables synthesizing clean, coherent articulated meshes, ensuring both geometric quality and functional properties. We believe \OURS{} will drive progress in 3D vision and embodied AI, opening new possibilities for digital agents to learn and interact with the physical world.

\myparagraph{Acknowledgements.}
This project is funded by the Bavarian State Ministry of Science and the Arts and coordinated by the Bavarian Research Institute for Digital Transformation (bidt), the ERC Starting Grant SpatialSem (101076253), and the German Research Foundation (DFG) Grant ``Learning How to Interact with Scenes through Part-Based Understanding". We thank Quan Meng, Yueh-Cheng Liu, and Haoxuan Li for the constructive discussions. We also thank Yueh-Cheng Liu for helping with the supplemental video.

{
    \small
    \bibliographystyle{ieeenat_fullname}
    \bibliography{main}
}

\clearpage
\setcounter{page}{1}
\maketitlesupplementary
\setcounter{section}{6}
\setcounter{figure}{6}
\setcounter{table}{5}
\setcounter{equation}{4}

\label{sec:appendix}

In this supplementary document, we provide additional details about \OURS{}.
In \cref{sec:method_details}, we give more implementation details of our method and loss functions.
We elaborate our data annotation process in \cref{sec:data_annotation}.
We also include additional quantitative comparisons in \cref{sec:additional_results}.
We encourage readers to watch the supplemental video to see more articulated object generations in action.

\section{Method Details}
\label{sec:method_details}

We use VQVAEs to model both part articulations and mesh geometries for our hierarchical transformers.
Our structure VQVAE encodes extra part-level features (e.g., semantics, geometry feature, and articulation joint), alongside vertex locations, into a compact latent space for articulation-aware structure generation, while our geometry VQVAE predicts additional junction face probabilities for coherent part mesh generation.

\subsection{Structure VQ-VAE}
The structure VQ-VAE encodes and quantizes features of bounding box triangles to learn a structured embedding space for articulated object structures. 
We construct a graph for the triangles by treating each triangle face as a node and connecting neighboring faces with undirected edges.
The input node features include positionally encoded triangle coordinates, face area, edge angles, and face normal vectors. These features are concatenated with part semantic, geometry, and articulation attributes projected onto the triangle nodes. The combined features are processed through 4 SAGEConv~\cite{hamilton2017inductive} graph convolutional layers, extracting a feature vector of dimension $768$ for each triangle.

At the bottleneck, these embeddings are quantized using a codebook of size 8192, enabling a compact representation of the structure. The decoder reconstructs triangle locations by predicting the logits of discretized coordinates, where both triangle and joint locations are mapped to a uniform grid of size $128^3$. 

Instead of directly predicting a discrete part semantic label, the structure VQ-VAE decoder regresses a continuous semantic feature vector from CLIP. The class label is then determined by computing the cosine similarity between the predicted feature vector and the CLIP features of a predefined set of part labels.

The decoder will output a set of joint information per triangle. To obtain a single set of joint predictions per part, the outputs are averaged across all triangles within the part.

This architecture effectively learns quantized embeddings for articulated object structures. These embeddings serve as the basis for the structure transformer, enabling the autoregressive generation of object structures with articulations.

\mypara{Loss Functions.}
As the triangle coordinates and joint locations are discretized, their reconstruction loss can be formulated as a cross-entropy loss:
\begin{equation}
    L_{recon} = \sum_{n=1}^{N}\sum_{k=1}^{128} \log\mathbf{P}_k,
\end{equation}
with $n$ being the face index and $\mathbf{P}_k$ representing the predicted probability distribution over the coordinate bins.
For part $i$, its semantic feature $\PartSemantics_i$ and geometry feature $\PartGeometry_i$ are supervised using $L_2$ regression loss:
\begin{equation}
    L_{regression} = ||\mathbf{y}_i - \hat{\mathbf{y}}_i||_2,
\end{equation}
with $y_i, \hat{y}_i$ being the ground truth and predicted feature vectors. 
\subsection{Structure Transformer}
We use a decoder-only transformer that has a standard GPT-2 architecture, \ie, 12 multi-headed self-attention layers, 12 heads, 768 as feature width, with a context length of 4608. The transformer is trained with cross-entropy loss for next-token index prediction.

\subsection{Geometry VQ-VAE}
The Geometry VQ-VAE encodes mesh triangle features using an architecture similar to the Structure VQ-VAE. Input triangle features, such as positional encoding, normals, and edge attributes, are processed through 4 SAGEConv~\cite{hamilton2017inductive} layers to extract feature embeddings of dimension $768$. These embeddings are quantized at the bottleneck using a vector quantization module with codebook size of 16384, enabling compact and efficient representation.

The 1D-ResNet decoder reconstructs the discretized triangle coordinates by minimizing a cross-entropy loss over a uniform grid. To enforce spatial and structural coherence between parts, the geometry decoder includes an additional channel that predicts the probability of each triangle being a junction triangle, i.e., triangles at the near boundary between adjacent parts. This prediction is supervised with a binary classification loss.

By incorporating junction triangle prediction, the Geometry VQ-VAE not only reconstructs accurate triangle meshes of the target part, but also learns the connectivity information cross parts, supporting smooth articulation and consistent geometry generation.

\begin{figure}[b]
    \centering
    \includegraphics[width=0.8\columnwidth]{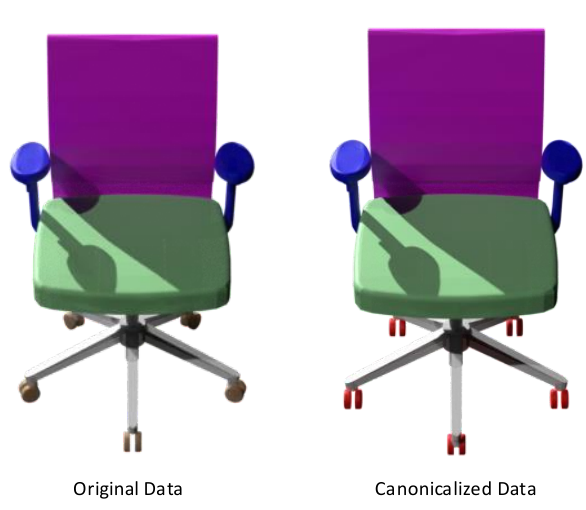}  
    \caption{We canonicalize the orientation of different articulated parts for consistent joint annotation.
    }
    \vspace{-0.5cm}
    \label{fig:wheel_anno}
\end{figure}

\begin{figure}[b]
    \centering
    \includegraphics[width=0.8\columnwidth]{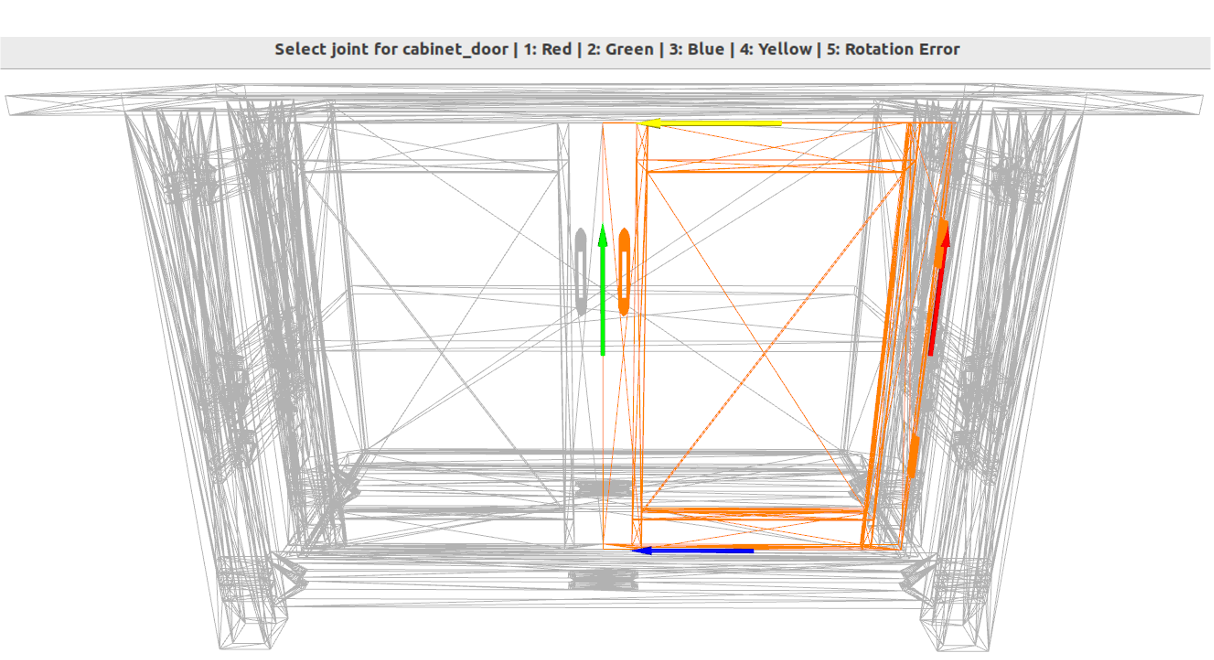}  
    \caption{Given a target part, our viewer visualizes the generated joint hypotheses for selection.
    }
    \vspace{-0.5cm}
    \label{fig:viewer}
\end{figure}

\section{Data Annotation}
\label{sec:data_annotation}
To effectively learn the distribution of articulated objects, we extend PartNet~\cite{mo2019partnet}, the largest dataset with object part annotations, by augmenting it with joint information. This augmentation significantly increases the diversity of articulated objects compared to the commonly used PartNet-Mobility~\cite{xiang2020sapien}.

\mypara{Part Canonicalization.}  
To ensure consistent and meaningful articulation properties, we canonicalize joint annotations. For prismatic joints, all locations are set to the origin of the object’s coordinate system. For revolute joints, we address inconsistencies in part orientations, for instance, chair wheels often have arbitrary orientations in the original dataset, resulting in misaligned revolute joints. To canonicalize these, we rotate each wheel around its vertical axis to align their orientations consistently, as shown in Fig.~\ref{fig:wheel_anno}.

\mypara{Joint Location Generation.}
For storage furniture and tables, revolute joints are typically located at the ``hinge'' of an articulated part, often corresponding to one of the four bounding box sides of the part. To automate this process, we generate four hypotheses for joint locations based on the bounding box configuration of the articulated part. An interactive viewer is then used to select the most reasonable joint location, as illustrated in Fig.~\ref{fig:viewer}.

{
\begin{table}[tp]
  \begin{center}
    \small
    \resizebox{\linewidth}{!}{
      \begin{tabular}{p{10mm}lrrrrr}
        \toprule
        Class               & \multicolumn{1}{c}{Method}                   & COV$\uparrow$  & MMD$\downarrow$ & 1-NNA          & FID$\downarrow$ & KID$\downarrow$   \\
        \midrule

        \multirow{4}{*}{Chair} & NAP~\cite{lei2023nap}                     &     20.9        &     5.4        &     97.3    &    212.6      &      0.207                              \\

                               & MeshGPT~\cite{siddiqui2024meshgpt}        &\textbf{34.5}&   4.2       &      \textbf{81.8}         &       24.0      &   \textbf{0.011}   \\

        \cmidrule{2-7}
                               & \OURS{}                                   &   27.3      & \textbf{3.8} & 85.8 & \textbf{23.8}      & 0.013   \\
        \bottomrule

        \multirow{4}{*}{Table} & NAP~\cite{lei2023nap}                     &     20.0    &    5.8      &      96.5         &       243.8     &   0.231    \\
        
                               & MeshGPT~\cite{siddiqui2024meshgpt}        & \textbf{43.0}&   2.9       &      \textbf{69.9}         &       \textbf{14.5}      &   \textbf{0.005}  \\
                               
        \cmidrule{2-7}
                               & \OURS{}                                   & 33.4          & \textbf{2.8}   & 77.9 & {15.1}   & {0.007}   \\
        \bottomrule

        \multirow{4}{*}{\shortstack{Storage\\Furniture}} & NAP~\cite{lei2023nap}     &    25.3  &    2.9    &      92.4      &    162.4     &    0.142   \\
         
                                                         & MeshGPT~\cite{siddiqui2024meshgpt}        &   38.7    &  2.3        &      81.6         &       9.3      &   \textbf{0.002}\\
                              
        \cmidrule{2-7}
                                                         & \OURS{}                    & \textbf{40.9} & \textbf{2.0}   & \textbf{78.7} & \textbf{8.6}   & {0.003}  \\
        \bottomrule

      \end{tabular}}
    \caption{Quantitative comparison on the task of unconditional mesh generation on a subset of categories from the PartNet~\cite{mo2019partnet} dataset. MMD values are multiplied by $10^3$. We evaluate the mesh quality at the resting state for all methods. We outperform the baselines in shape quality, visuals, and compactness metrics.}
    \label{tab:quantative_evaluation_resting}
    \vspace{-8mm}
  \end{center}

\end{table}
}

\begin{figure}[t]
    \centering
    \includegraphics[width=0.95\columnwidth]{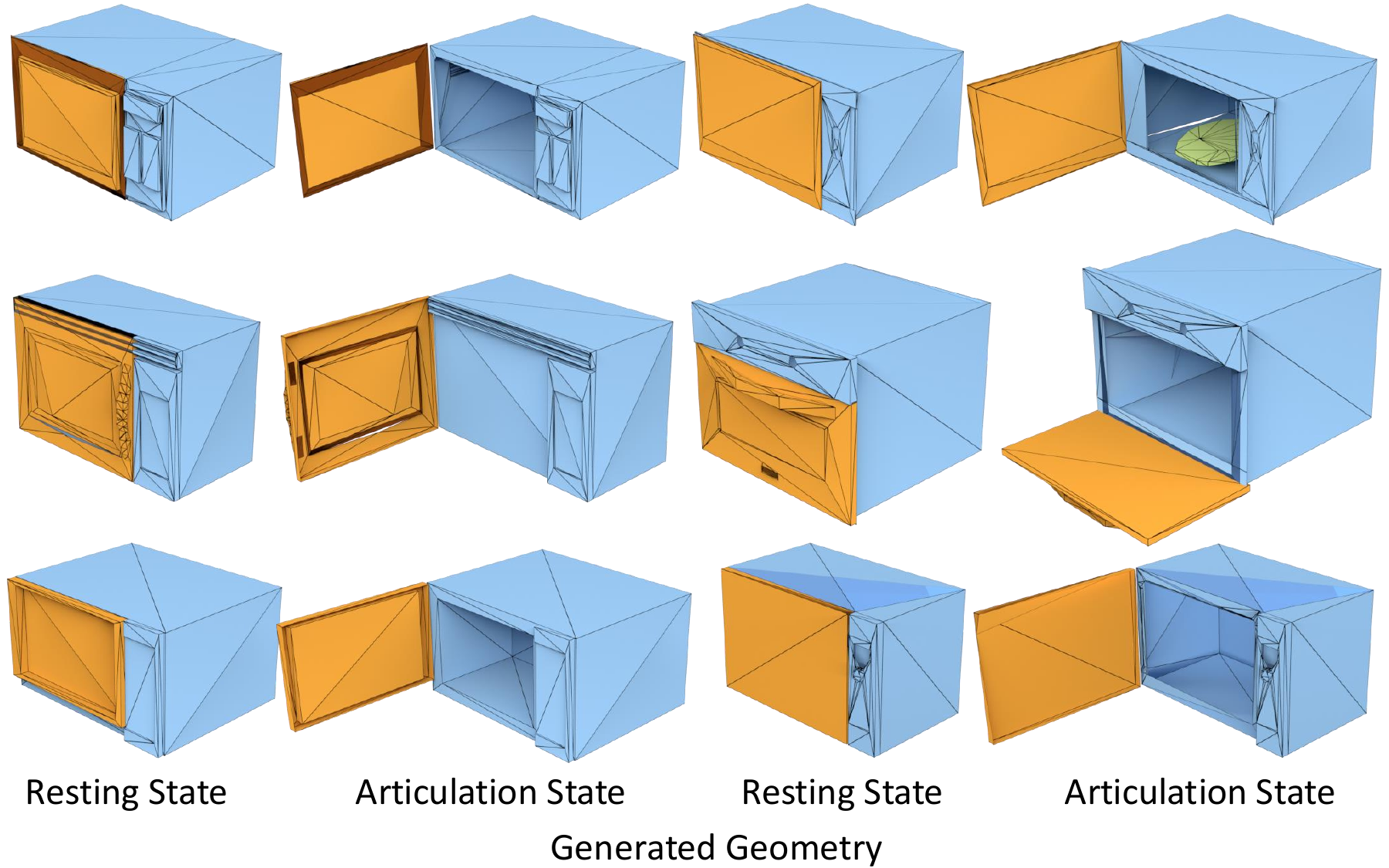}  
    \caption{Our method can generate sharp geometry and realistic articulations for microwaves.}
    \vspace{-0.3cm}
    \label{fig:microwave}
\end{figure}

\begin{figure*}[t]
    \vspace{-0.3cm}
    \centering
    \includegraphics[width=0.95\textwidth]{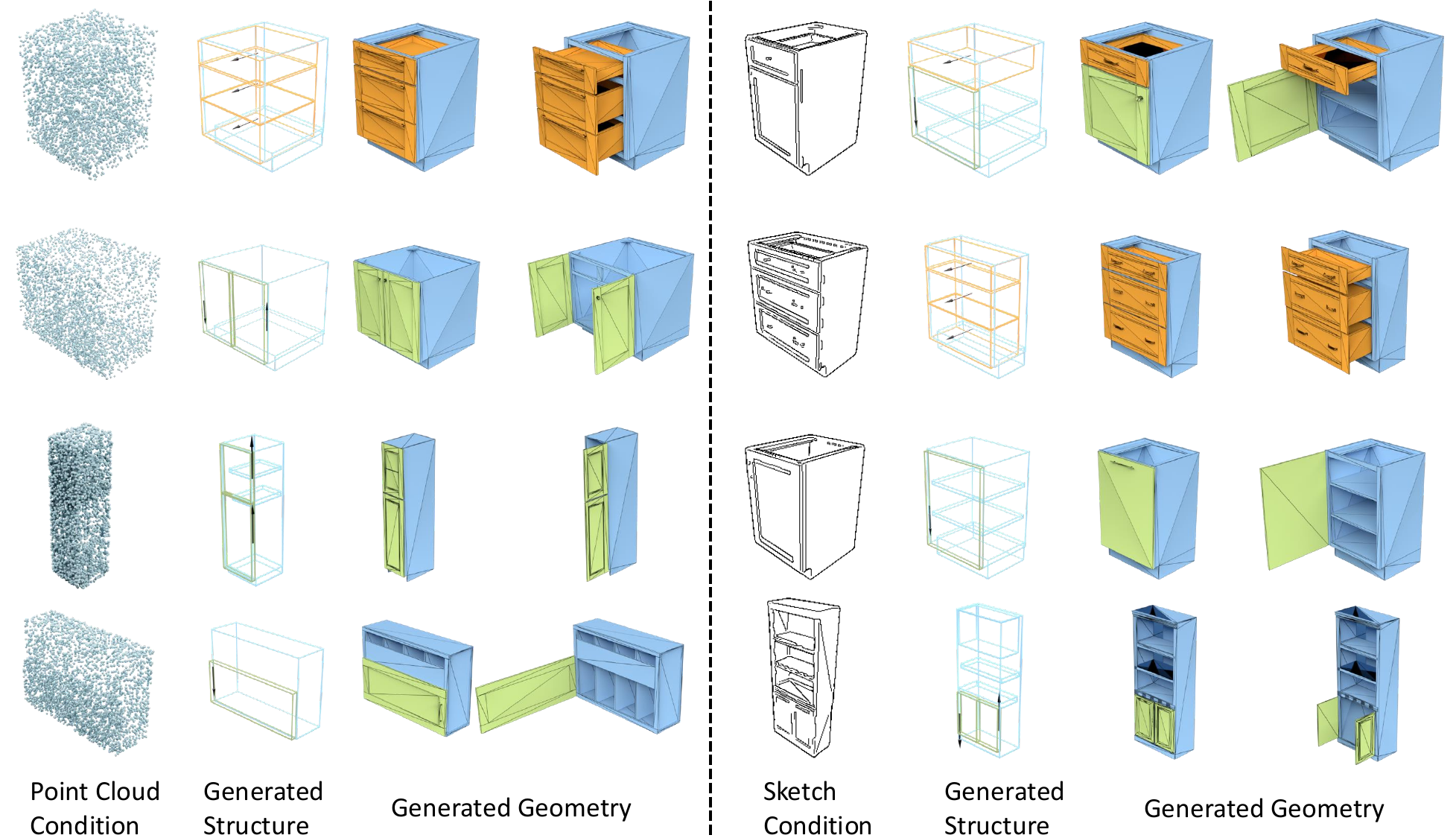}  
    \caption{Conditional Generation. Articulated structures and geometry are generated conditioned on point clouds or sketches.
    }
    \vspace{-0.3cm}
    \label{fig:condgen_supp}
\end{figure*}

\mypara{Joint Verification.}
To validate the joint annotations, we render the object at various articulation states and visually inspect the plausibility of the generated motions. This step ensures the accuracy of joint locations and their associated articulation properties, providing high-quality annotations for articulated objects.

\section{Additional Results}
\label{sec:additional_results}

\mypara{Quantitative Comparison at Resting State.}
We compare the mesh generation quality of our method with NAP~\cite{lei2023nap}, and the state-of-the-art direct mesh generation approach, MeshGPT~\cite{siddiqui2024meshgpt}. Since MeshGPT does not predict object part and articulation information, the evaluation is performed on generated meshes in their canonical resting state. As shown in Tab.~\ref{tab:quantative_evaluation_resting}, we also achieve comparable COV scores to MeshGPT while outperforming the MMD score, indicating higher fidelity in the generated shapes.
Notably, our method shows significant mesh generation improvement over NAP on all metrics.

\mypara{Additional Categories}.
We show additional results on microwaves in Fig.~\ref{fig:microwave}.

\begin{figure}[t]
    \centering
    \vspace{-0.3cm}
    \includegraphics[width=0.95\columnwidth]{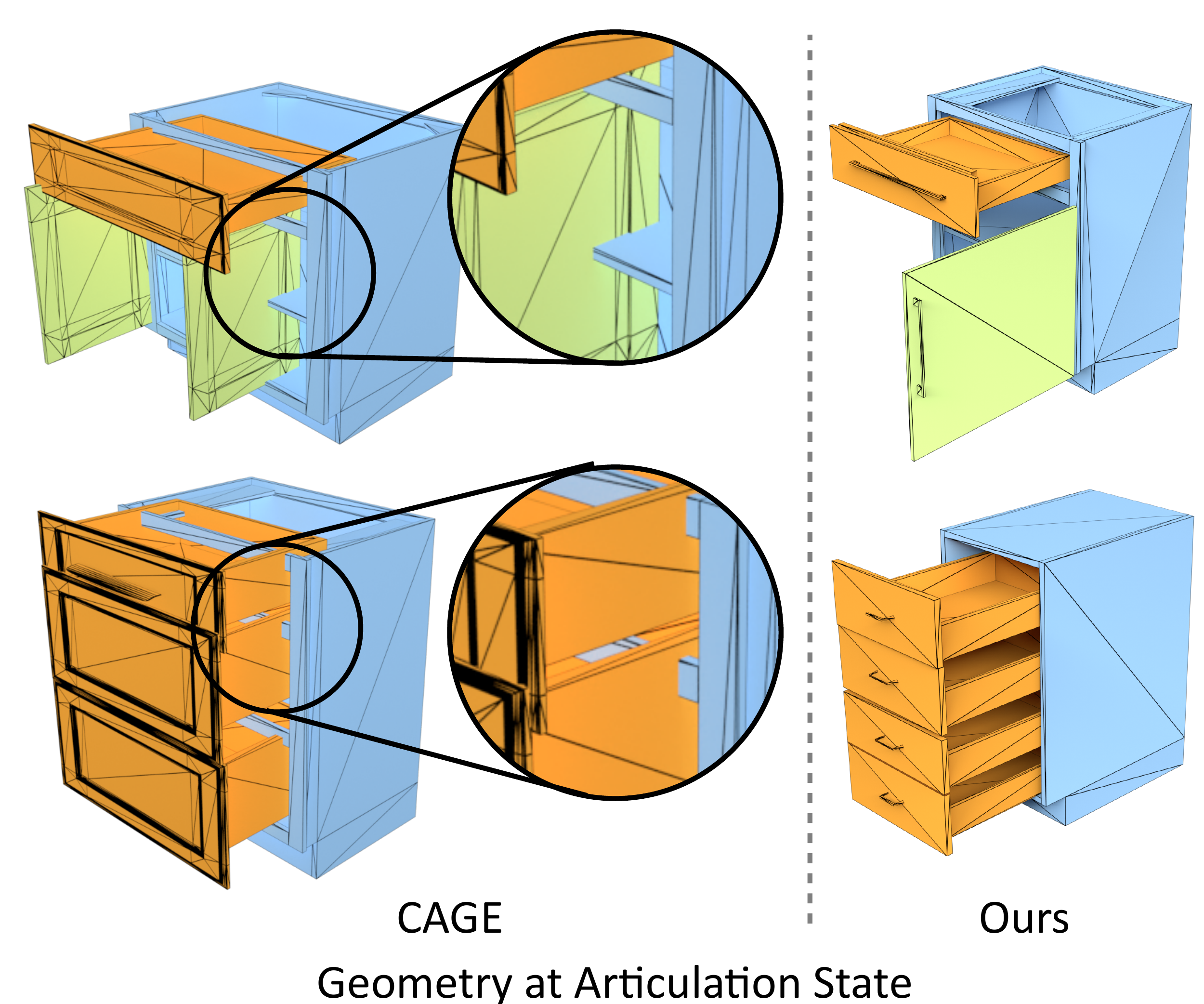}  
    \caption{Visual comparison with CAGE. Our method can generate coherent shapes with sharp geometry.
    }
    \vspace{-0.3cm}
    \label{fig:cage}
\end{figure}

\mypara{Qualitative Comparison with CAGE.}
We compare \OURS{} with CAGE~\cite{liu2024cage}, which retrieves part geometry based on conditionally generated articulated structures. As shown in Fig.~\ref{fig:cage}, our method achieves coherent shape synthesis while achieving sharp geometry details, avoiding undesired part collision and inconsistencies between parts.

\mypara{Conditional Generation}.
\OURS{} can generate articulated structure and geometry conditioned on point clouds or sketch images. We extract input features using Michelangelo~\cite{zhao2024michelangelo} for 3D point clouds and Radio~\cite{ranzinger2024radio} for sketches. A linear layer projects these features to match the structure transformer $\StructureTransformer$'s feature space, appending them to the beginning of the structure sequence. The transformer $\StructureTransformer$ then learns to generate articulation-aware structure bounding boxes. Since the structure codebook $\StructureCodebook$ remains fixed, the geometry transformer $\GeometryTransformer$ requires no finetuning. As shown in Fig.~\ref{fig:condgen_supp}, given the 3D/2D conditions, our method can generate plausible articulated structures, which then guide faithful synthesis of part geometry.

\end{document}